%% file: main.tex
\documentclass{article}

\usepackage{arxiv}

\usepackage[utf8]{inputenc} % allow utf-8 input
\usepackage[T1]{fontenc}    % use 8-bit T1 fonts
\usepackage{hyperref}       % hyperlinks
\usepackage{url}            % simple URL typesetting
\usepackage{booktabs}       % professional-quality tables
\usepackage{amsfonts}       % blackboard math symbols
\usepackage{nicefrac}       % compact symbols for 1/2, etc.
\usepackage{microtype}      % microtypography
\usepackage{graphicx}
\usepackage{natbib}
\usepackage{doi}

\usepackage{algorithm}%
\usepackage{algorithmicx}%
\usepackage{algpseudocode}%
\usepackage{multirow}
\usepackage{tabularx}
\usepackage{mathtools}
\usepackage{booktabs}
\usepackage{color}
\usepackage{makecell}
\usepackage{placeins}
\usepackage[figuresright]{rotating}%

\DeclarePairedDelimiter\floor{\lfloor}{\rfloor}
\DeclarePairedDelimiter\ceil{\lceil}{\rceil}

\newtheorem{definition}{Definition}
\newtheorem{proposition}{Proposition}

\title{Interpretable Decision Trees Through MaxSAT}
\author{
    Josep Alòs \\
    Universitat de Lleida \\
    \texttt{josep.alos@udl.cat}
    \And
    Carlos Ansotegui\\
    Universitat de Lleida \\
    \texttt{carlos.ansotegui@udl.cat}
    \And
    Eduard Torres\\
    Universitat de Lleida \\
    \texttt{eduard.torres@udl.cat}
}
\date{July 2022}

\renewcommand{\headeright}{}
\renewcommand{\undertitle}{}
\renewcommand{\shorttitle}{Interpretable Decision Trees Through MaxSAT}

\hypersetup{
pdftitle={Interpretable Decision Trees Through MaxSAT},
pdfsubject={cs},
pdfauthor={Josep Alòs, Carlos Ansótegui, Eduard Torres},
pdfkeywords={Decision trees, MaxSAT, Explainable AI},
}

\begin{document}
\maketitle
\begin{abstract}
We present an approach to improve the accuracy-interpretability trade-off of Machine Learning (ML) Decision Trees (DTs). In particular, we apply Maximum Satisfiability technology to compute Minimum Pure DTs (MPDTs). We improve the runtime of previous approaches and, show that these MPDTs can outperform the accuracy of DTs generated with the ML framework \emph{sklearn}.
\end{abstract}


%%%%%%%%%%%%%%%%%%%%%%%%%%%%%%%%%%%%%%%%%%%%%%%%%%%%%%
\section{Introduction}\label{sec1}

Recently, there has been a growing interest in creating synergies between Combinatorial Optimization (CO) and Machine Learning (ML), and vice-versa. This is a natural connection since ML algorithms can be seen in essence as optimization algorithms that try to minimize prediction error. In this paper, we focus on how CO techniques can be applied to improve the interpretability and accuracy of decision tree classifiers in ML.

A Decision Tree (DT) classifier is a supervised ML technique that builds a tree-structured classifier, where internal nodes represent the features of a dataset, branches represent the decision rules and each leaf node represents the outcome. In essence, every path from the root to a leaf is a classification rule that determines to which class belongs the input query.

ML research has been traditionally concerned with developing accurate classifiers, i.e., classifiers whose prediction error is as low as possible. However, lately, the research community is looking at the trade-off between accuracy and interpretability, where the latter refers to the human ability to understand how the classifier works. Interpretability is fundamental in many senses: to explain ML models and understand their value and accuracy, to debug machine learning algorithms, to make more informed decisions, etc. 

Among supervised learning techniques, decision tree classifiers are interpretable in itself, as long as they are \emph{small}, i.e., as few rules as possible, and as short as possible. As stated in \cite{DBLP:conf/cp/BessiereHO09}, informally, this follows from the principle of Occam’s Razor that states that one should prefer simpler hypotheses. However, notice that this is just one of the diverse aspects that may have to do with interpretability at least from a human perspective. 

CO approaches have been applied to compute \emph{small} trees in the sense of building decision trees of minimum size (number of nodes) or minimum depth. Minimizing the size corresponds indirectly to minimizing the number of rules associated with the DT and minimizing the depth corresponds to minimizing the length of the rules. Among these approaches we find several contributions coming from Operation Research (OR)~\cite{DBLP:conf/aaai/AglinNS20},  \cite{verwerLearningOptimalClassification2019} and from Constraint Programming (CP)~\cite{DBLP:conf/nips/HuRS19},  \cite{verhaegheLearningOptimalDecision2019}. 

Another question is whether these DTs are pure (i.e., classify correctly the training dataset). To avoid overfitting on the training set, and therefore get lower accuracy on the test set, there is a rule of thumb that makes us prefer some impurity to get more generalization power. However, it is difficult to anticipate how much impurity we should allow.

In this paper, we first address the question of how to efficiently compute minimum pure DTs and discuss later how accurate they are.
We focus on CP techniques where SATisfiability (SAT)~\cite{handbooksat} based approaches have been shown to be very effective. Particularly, we will work on the Maximum Satisfiability (MaxSAT) problem, which is the optimization version of the SAT problem.

We revisit the SAT approach in \cite{DBLP:conf/ijcai/NarodytskaIPM18} which encodes into SAT the decision query of whether there is a pure DT  of exactly $n$ nodes. Then, by performing a sequence of such SAT queries the optimum can be located. We extend this SAT approach to the MaxSAT case, i.e., we encode into MaxSAT the query representing which is the minimum number of nodes $n$ for which such a pure DT exists. 

Equipped with a MaxSAT encoding we can now leverage the power of current (and \emph{future}) MaxSAT solvers. Notice that the MaxSAT formalism allows us to state explicitly that we are dealing with an optimization problem, while the original incremental SAT solving approach in \cite{DBLP:conf/ijcai/NarodytskaIPM18} is agnostic in this sense and the objective has to be managed by an \textit{ad hoc} piece of code. Our experimental results show that our approach in practice is able to obtain minimal pure DTs on a wide variety of benchmarks faster than the original SAT incremental counterpart in \cite{DBLP:conf/ijcai/NarodytskaIPM18} even using the same underlying SAT solver. We also show that our approach provides comparable performance to other approaches~\cite{DBLP:conf/sat/JanotaM20} that exploit explicitly structural properties of the DT.

Secondly, and the main contribution of the paper, we show that by exploring the space of optimal solutions (minimal pure DTs) we can be indeed competitive with respect to DT classifiers generated with sklearn in terms of producing smaller trees but yet accurate enough.

In order to achieve the main goal of this paper, we solve yet another optimization problem that extracts multiple optimal solutions, maximizing their \emph{diversity}, from the MaxSAT encoding that allows us to compute minimum pure DTs for the training set. Then, thanks to a validation process, we select among these multiple diverse solutions one that is expected to report better accuracy on the test set.

This is an interesting result since all the optimal solutions we consider are minimum pure DTs for the training set, yet they do not seem to suffer substantially from the \emph{overfitting} phenomena. Notice that what makes them special is that they all are of minimum size, a feature that can not be exploited by current DT ML approaches.

Finding MPDTs is an NP-hard problem, and therefore a time-consuming process. Our goal is to provide a solution for applications where
high interpretability and accuracy of DTs prevail over the training speed, such as the AI systems considered as ``high-risk'' in the proposed regulation~\cite{european_commission2021}. In particular, this solution is suitable for systems where re-training the model frequently is not desired, to avoid the opacity induced by such frequent changes in the model.
This would be the case of models used in administration selection processes, financial risk ratings, or patient screening.

%%%%%%%%%%%%%%%%%%%%%%%%%%%%%%%%%%%%%%%%%%%%%%%%%%%%%%
%%%%%%%%%%%%%%%%%%%%%%%%%%%%%%%%%%%%%%%%%%%%%%%%%%%%%%

\section{Preliminaries} \label{sec:preliminaries}

\begin{definition} An example $e$ is a set of pairs $\langle f_r,v \rangle$ plus the class label $c$, where $f_r$ is a feature and $v$ is the value assigned to the feature.  A Decision Tree (DT) is a set of rules (namely paths) that are constructed by traversing a tree-like structure composed of questions (decision nodes) and answers until a terminal node with a label (leaf) is found. The depth of a node is the length of the path (i.e. count of nodes traversed) from the root to this node. The depth of the DT is the length of the largest path from the root to any leaf. A Complete Decision Tree (CDT) is a tree where all the leaves are located at the same depth. A Pure Decision Tree (PDT) for a set of examples $\varepsilon$, $PDT(\varepsilon)$, is a DT that classifies correctly all the examples \cite{DBLP:journals/bioinformatics/HautaniemiKIWL05}. 
By definition, its accuracy on this set is $100\%$. A Minimum Pure Decision Tree (MPDT) for a set of examples $\varepsilon$, $MPDT(\varepsilon)$, is a DT that is proven to be the smallest PDT for $\varepsilon$ in terms of size (i.e. the number of nodes in the DT). We denote the minimum size as $\lvert MPDT(\varepsilon) \rvert$.

\end{definition}

\begin{definition} A literal is a propositional variable~$x$ or a negated propositional variable~$\neg x$. A clause is a disjunction of literals. A Conjunctive Normal Form (CNF) is a conjunction of clauses. A weighted clause is a pair $(c,w)$, where $c$ is a clause and $w$, its weight, is a natural number or infinity. A clause is hard if its weight is infinity (or no weight is given); otherwise, it is soft. A Weighted Partial MaxSAT instance is a multiset of weighted clauses.
\end{definition}

\begin{definition} A truth assignment for an instance $\phi$ is a mapping that assigns to each propositional variable in $\phi$ either 0 (False) or 1 (True). A truth assignment is \emph{partial} if the mapping is not defined for all the propositional variables in $\phi$. A truth assignment $I$ satisfies a literal $x$ $(\neg x)$ if $I$ maps $x$ to 1 (0); otherwise, it is falsified. A truth assignment $I$ satisfies a clause if $I$ satisfies at least one of its literals; otherwise, it is violated or falsified. The cost of a clause $(c,w)$ under $I$ is 0 if $I$ satisfies the clause; otherwise, it is $w$.
Given a partial truth assignment $I$, a literal or a clause is undefined if it is neither satisfied nor falsified.  A clause $c$ is a unit clause under $I$ if $c$ is not satisfied by $I$ and contains exactly one undefined literal.  The cost of a formula $\phi$ under a truth assignment $I$, $cost(I, \phi)$, is the aggregated cost of all its clauses under $I$.
\end{definition}

\begin{definition}
The Weighted Partial MaxSAT problem for an instance $\phi$ is to find an assignment in which the sum of weights of the falsified soft clauses, cost($\phi$), is minimal and, the hard clauses are satisfied. The Partial MaxSAT problem is the Weighted Partial MaxSAT problem where all weights of soft clauses are equal. The SAT problem is the Partial MaxSAT problem when there are no soft clauses. An instance of Weighted Partial MaxSAT, or any of its variants, is unsatisfiable if its optimal cost is $\infty$. A SAT instance $\phi$ is satisfiable if there is a truth assignment $I$, called model, such that $cost(I, \phi) = 0$.
\end{definition}

\begin{definition} 
A \emph{pseudo-Boolean} constraint (PB) is a Boolean function of the  form $ \sum_{i=1}^{n} q_il_i \diamond k $, where $k$ and the $q_i$ are integer constants, $l_i$ are literals, and $\diamond \in \{<, \leq, =, \geq, >\}$ is a comparison operator. An \emph{At Least K} (AtLeastK) constraint  is a PB of the form $ \sum_{i=1}^{n} l_i \ge k $. An \emph{At Most K} (AtMostK) constraint is a PB of the form $ \sum_{i=1}^{n} l_i \le k $.
\end{definition}

%%%%%%%%%%%%%%%%%%%%%%%%%%%%%%%%%%%%%%%%%%%%%%%%%%%%%%
%%%%%%%%%%%%%%%%%%%%%%%%%%%%%%%%%%%%%%%%%%%%%%%%%%%%%%

\section{Related Work}\label{sec:related_work}

In a pioneering work, \cite{DBLP:conf/cp/BessiereHO09}, the first SAT approach to compute MPDTs was presented.  They encode into SAT the decision query of whether there exists a PDT of size $n$. They use as upper bound the solution from a CP model. Then, they perform a sequence of queries to a SAT solver (in the SAT4J package \cite{LeBerre2010}) where $n$ is iteratively bounded by adding an \emph{AtLeastK} constraint on the number of \emph{useless} nodes that must be in the solution.

In \cite{DBLP:conf/ijcai/NarodytskaIPM18} a more efficient SAT encoding was provided to encode the decision query of whether there exists a PDT of size $n$. The authors also solve the problem through a sequence of SAT queries.
In this case, the upper bound is computed with the algorithm ITI \cite{DBLP:journals/ml/Utgoff89}. Then, at every iteration, they reencode the problem with the next possible value of $n$. We refer to this approach as \emph{MinDT}.

In \cite{DBLP:conf/aaai/Avellaneda20} two methods are proposed. The \textit{Infer-depth} approach iteratively decreases the depth of the tree, till it locates the minimum depth to build a complete PDT. The \textit{Infer-size} approach computes the MPDT with the depth found with \textit{Infer-depth}.
While both approaches compute \emph{small} PDTs, none of them guarantees to find MPDTs.

In \cite{DBLP:conf/sat/JanotaM20},  one of the approaches, \emph{dtfinder}, split the search space into the possible \emph{topologies} (tree layouts) of the DT till a certain depth. At least one of these \emph{topologies} is guaranteed to belong to an MPDT.

In \cite{DBLP:conf/ijcai/Hu0HH20}, \emph{DT(max)}, the approach answers to which is the best tree in terms of accuracy in the training set that can be obtained given an upper bound on the depth. Note that if the upper bound is smaller than the minimal depth of MPDT, their approach reports a tree with less than 100\% of accuracy on the training set (i.e. impure tree). Additionally, an AdaBoost \cite{FREUND1997119} approach based on MaxSAT is presented. We will not explore AdaBoost strategies in this paper, since they produce ensembles, which are composed of several DTs. Having more than one DT weakens the interpretability of the classifier.

Table~\ref{tab:summary_sota} shows a summary, presenting the objective function, whether the approach returns a PDT and if it is minimum (MPDT). The only algorithms that are complete in the sense of certifying the minimum size of the PDTs are \cite{DBLP:conf/cp/BessiereHO09}, MinDT \cite{DBLP:conf/ijcai/NarodytskaIPM18} and \emph{dtfinder} \cite{DBLP:conf/sat/JanotaM20}. 

We will focus on comparing our approach \emph{MPDT} (see Section~\ref{sec:mindt_ms}) with \emph{MinDT} in terms of computing efficiently MPDTs since other approaches build on \emph{MinDT}. In terms of computing rather \emph{small} trees with good accuracy, we will compare with \emph{DT(max)} \cite{DBLP:conf/ijcai/Hu0HH20} which in contrast to the other approaches focuses also on maximizing the accuracy.

\begin{table}[ht]
    \centering
    \begin{tabular}{l|l|r|r}
        \toprule
        Algorithm                & Objective  & PDT & MPDT \\
        \midrule
        \textit{Bessiere et al.} & min. size  & yes & yes \\
        \emph{MinDT}                    & min. size  & yes & yes \\
        \emph{Infer-depth}              & min. depth & yes & no \\
        \emph{Infer-size}               & min. depth, size & yes & no \\
        \emph{dtfinder}           & min. size & yes & yes \\
        \emph{DT(max)}                  & max. acc & no & no \\
        MPDT                     & min. size & yes & yes \\
        \bottomrule
    \end{tabular}
    \caption{State-of-the-art: SAT-based approaches for computing DTs}
    \label{tab:summary_sota}
\end{table}

%%%%%%%%%%%%%%%%%%%%%%%%%%%%%%%%%%%%%%%%%%%%%%%%%%%%%%
%%%%%%%%%%%%%%%%%%%%%%%%%%%%%%%%%%%%%%%%%%%%%%%%%%%%%%

\section{Computing MPDTs with MaxSAT}\label{sec:mindt_ms}

We extend the SAT encoding in \cite{DBLP:conf/ijcai/NarodytskaIPM18}, MinDT, to a Partial MaxSAT (and its weighted variant) encoding that queries for an MPDT given an upper bound in size of $n$. Our approach is inspired by the MaxSAT encoding in \cite{Ansotegui2013} for constructing minimum covering arrays. Recent work in \cite{DBLP:conf/ijcai/Hu0HH20} follows a similar approach with the $m_i$ variable, which is used to relax all the constraints of the encoding. We selectively use the relevant variable only in the constraints that require it. For completeness and the aim of reproducibility, we describe in full detail the original SAT encoding and the extensions to the MaxSAT case, highlighting the changes from the original encoding.

We assume that: (1) the features are binary; (2) the root node is labeled as node 1 (the rest of them are labeled into breadth-first order); and (3) positive (negative) answers to decision nodes are assigned to the right (left) child. In Table~\ref{tab:symbols_and_functions} and Table~\ref{tab:sat_variables} we show the symbols, functions, and the propositional variables of the encoding.

\begin{table}[ht]
    {
    \centering
    \begin{tabularx}{\columnwidth}{l|l}
        \toprule
        Sym/fun              & Definition \\
        \midrule
        $\varepsilon$        & The set of observations used to train the tree\\
        $\varepsilon^{+(-)}$ & The set of positive (negative) observations\\
        $n$                  & The upper bound on the number of nodes \\
        $N$                  & Range of nodes, starting at 1 \\
        $N^e$                & Range of even-indexed nodes \\
        $N^o$                & Range of odd-indexed nodes \\
        $k$                  & Number of features for each observation, starting at 0 \\
        $F$                  & Range of features for each observation, starting at 0 \\
        $lr(i)$              & Set of potential left child nodes for node $i$ \\
        $rr(i)$              & Set of potential right child nodes for node $i$\\
        $\sigma(r,q)$        & The value of the feature $f_r$ at observation $e_q$\\
        $\eta(i)$            & Related variable $\eta_{i'}$ for the node $i$ \\
        \bottomrule
    \end{tabularx}
    }
    \caption{Symbols and functions used in the MaxSAT encoding}
    \label{tab:symbols_and_functions}
\end{table}

\begin{table*}[ht]
    \centering
    \begin{minipage}{\textwidth} \centering
    \begin{tabularx}{\textwidth}{l|l|X}
        \toprule
        \multicolumn{1}{c|}{Variable} & \multicolumn{1}{c|}{Index's Range} & \multicolumn{1}{c}{Represents}  \\
        \midrule
        $v_i$       & $i \in N$                 & True iff. node $i$ is a leaf \\
        $l_{i,j}$   & $i \in N, j \in lr(i)$    & True iff. node $j$ is the left child of node $i$\\
        $r_{i,j}$   & $i \in N, j \in rr(i)$    & True iff. node $j$ is the right child of node $i$ \\
        $p_{j,i}$   & $i \in N, j \in N$        & True iff. node $i$ is the parent of node $j$ \\
        $\eta_i$    & $i \in N^o$               & True iff. nodes $i$ and $i-1$ are used \\
        \hline
        $a_{r,j}$   & $r \in F, j \in N$        & True iff. feature $f_r$ is assigned to node $j$ \\
        $u_{r,j}$   & $r \in F, j \in N$        & True iff. feature $f_r$ has been discriminated in the path between the root and node $j$ \\
        $d^v_{r,j}$ & $v \in \{0, 1\}, r \in F, j \in N$ & True iff. feature $f_r$ is discriminated for value $v$ by node $j$ or one of its ancestors \\
        $c_j$       & $j \in N$                 & True iff. class assigned to node $j$ is 1\\
        \hline
        $\lambda_{t,i}$ & $i \in N, t \in [0, \ceil{\frac{i}{2}}]$ \footnote{ In \cite{DBLP:conf/ijcai/NarodytskaIPM18} it is stated that  $ t \leq \floor{\frac{i}{2}}$. This does not hold for the last node of a complete tree.} & True iff. there are (at least) $t$ leaves until (and included) node $i$ \\
        $\tau_{t,i}$ & $i \in N, t \in [0, i]$ & True iff. there are (at least) $t$ decision nodes until (and included) node $i$ \\
        \bottomrule
     \end{tabularx}
     \end{minipage}
     \caption{Variables used in the MaxSAT encoding}
     \label{tab:sat_variables}
\end{table*}

We add a new variable $\eta_i$  which represents if a node $i$ is used. The encoding works on an upper-bounded number of nodes. If the upper bound is not optimal, some nodes will not be used.
Since nodes are numbered in breadth-first order, two consecutive nodes with the same parent will be both in the solution or not. Therefore, we define function $\eta(i)$ as:

\begin{equation*}
    \color{red}{}
    \eta(i) = 
    \begin{cases}
        \eta_{i+1} & \mathrm{if}\; i \in N^e; \\
        \eta_{i} & \mathrm{if}\; i \in N^o
    \end{cases}
\end{equation*}

The soft clauses in the Partial MaxSAT encoding  that describe the objective function of the MPDT problem are: 
\begin{equation}\label{SOFTN}\tag{$SoftN$}
\color{red}{}
\bigwedge_{i \in N^{o}}(\neg \eta_i, 1)
\end{equation}

Now, we describe the hard constraints.
To encode PB constraints we use OptiLog \cite{optilog2021}.\\[1mm]

\textbf{Node usage hard clauses:}

\noindent\textbullet\ If an odd node is not used, then both himself and the previous one are not leaves.
\begin{equation}\label{eq:maxsat_leaves_are_used}
    \color{red}{
    \bigwedge_{i \in N^o\ \lvert \ i\ > 1}
    \neg \eta_i \rightarrow \neg v_i \land \neg v_{i-1}
    }
\end{equation}

\noindent\textbullet\ If a node is used, the previous ones are used:
\begin{equation}\label{eq:maxsat_used_consecutive}
    \color{red}{
    \bigwedge_{i \in N^o\ \lvert \ i\ > 1}
    \eta_i \rightarrow \eta_{i-2}
    }
\end{equation}

\noindent\textbullet\ To enforce a lower bound  $lb$ (3 by default) we add the following clause. 

\begin{equation}\label{eq:maxsat_first_three_used}
    \color{red}{\eta(lb)}
\end{equation}

\textbf{Tree layout hard clauses:}

\noindent\textbullet\ The root node is not a leaf.
\begin{equation}\label{eq:c1_root_no_leaf}
    \neg v_1
\end{equation}

\noindent\textbullet\ (i) Leaves do not have a left child, and (ii) if node $i$ is used in the DT, and has no left child candidates, then it is a leaf.
\begin{equation}\label{eq:c2_leaf_no_children}
    \bigwedge_{i \in N}
    \begin{cases}
        \bigwedge_{j \in lr(i)} (v_i \rightarrow \neg l_{i,j})) & \color{red}{\mathrm{if}\ \vert lr(i)\vert \neq 0}\\
        \color{red}{\eta(i) \rightarrow v_i} & \color{red}{\mathrm{if}\ \vert lr(i)\vert = 0}
    \end{cases}
\end{equation}

\noindent\textbullet\ Node $j$ is the left child of node $i$ iff node $j+1$ is the right child of node $i$.
\begin{equation}\label{eq:c3_left_right_consecutive}
    \bigwedge_{i \in N}
    \bigwedge_{j \in lr(i)} (l_{i,j} \leftrightarrow r_{i,j+1})
\end{equation}

\noindent\textbullet\ If a used node $i$ is not a leaf, then it has exactly one left child.
\begin{equation}\label{eq:c4_decision_one_left_child}
    \bigwedge_{i \in N}
    \left(
        (\neg v_i \color{red}{\land \eta(i)}\color{black}{})
        \rightarrow
        \left(
            \sum_{j \in lr(i)} l_{i,j} = 1
        \right)
    \right)
\end{equation}

\noindent\textbullet\ The symmetry equations on $p_{j,i},l_{i,j}$ and $r_{i,j}$.
\begin{align}\label{eq:c5_parent_to_left_and_right}
    \begin{split}
        \bigwedge_{i \in N}
        \bigwedge_{j \in lr(i)}
        (p_{j,i} \leftrightarrow l_{i,j}) \\
        \bigwedge_{i \in N}
        \bigwedge_{j \in rr(i)}
        (p_{j,i} \leftrightarrow r_{i,j})
    \end{split}
\end{align}

Given a pair of $i,j$ nodes, we have that $j \in lr(i) \cup rr(i)$, and $lr(i) \cap rr(i) = \emptyset$. This, combined with Constraint~\ref{eq:c5_parent_to_left_and_right}, sets an equivalence between the variable $p_{j,i}$ and one of $l_{i,j}$ or $r_{i,j}$ depending on the value $j$ being even or odd. Thus, in practice, we can perform a direct substitution of the variable $p_{j,i}$ with the corresponding one of those two variables in the encoding, and remove the $p_{j,i}$ variable and the Constraint~\ref{eq:c5_parent_to_left_and_right} altogether. In this description, for clarity purposes, we kept the $p_{j,i}$ variable in the encoding.

\noindent\textbullet\ If node $j$ is used in the DT (and it is not the root) it will have exactly one parent.
\begin{equation}\label{eq:c6_nodes_have_parent}
    \bigwedge_{j \in N\ \lvert \ j > 1}
    \left(
        \color{red}{\eta(j) \rightarrow}\color{black}{}
        \sum_{i=\floor{\frac{j}{2}}}^{j-1} p_{j,i} = 1
    \right)
\end{equation}

\textbf{Feature assignment hard clauses:}
\noindent\textbullet\ Feature $f_r$ with value $0$ (1) is already discriminated by node $j$ iff the feature $f_r$ is attached to the parent of $j$ and node $j$ is the right (left) child or the feature $f_r$ with value $0$ (1) is already discriminated by the parent of $j$.
\begin{align}
    \bigwedge_{j \in N, r \in F}
    \left(
        d_{r,j}^{0} \leftrightarrow    
        \bigvee\limits_{i=\floor{\frac{j}{2}}}^{j-1}
        \left(
            (a_{r,i} \land r_{i,j})
            \lor
            (p_{j,i} \land d_{r,i}^{0})
        \right)
    \right) \\
    \bigwedge_{j \in N, r \in F}
    \left(
        d_{r,j}^{1} \leftrightarrow
        \bigvee\limits_{i=\floor{\frac{j}{2}}}^{j-1}
        \left(
            (a_{r,i} \land l_{i,j})
            \lor
            (p_{j,i} \land d_{r,i}^{1})
        \right)
    \right)
\end{align}

\noindent\textbullet\ No feature is discriminated at the root node.
\begin{equation}\label{eq:c7_root_does_not_discriminate}
    \bigwedge_{r \in F} \left(
    \neg d_{r,1}^{0} \land \neg d_{r,1}^{1}
    \right)
\end{equation}

\noindent\textbullet\ If some feature $f_r$ is already discriminated by node $i$, then the feature can not be assigned to any of its children.
\begin{equation}\label{eq:c8_feature_only_once}
    \bigwedge_{j \in N, r \in F}
    \bigwedge\limits_{i=\floor{\frac{j}{2}}}^{j-1} ((u_{r,i} \land p_{j,i}) 
    \rightarrow
    \neg a_{r,j})
\end{equation}

\noindent\textbullet\ Feature $f_r$ is discriminated at node $j$ iff $f_r$ has been assigned to node $j$ or it is already discriminated at the parent of $j$.
\begin{equation}\label{eq:c9_feature_discrimination}
    \bigwedge_{j \in N, r \in F}
    \left(
        u_{r,j}
        \leftrightarrow
        \left( a_{r,j} \lor \bigvee\limits_{i=\floor{\frac{j}{2}}}^{j-1} (u_{r,i} \land p_{j,i})\right)
    \right)
\end{equation}

\noindent\textbullet\ If node $j$ is not a leaf and it is in the DT, then it has exactly one feature assigned to it.
\begin{equation}\label{eq:c10_decision_one_feat}
    \bigwedge_{j \in N}
    \left(
    (\neg v_j \color{red}{\land \eta(j)}\color{black}{})
    \rightarrow
    \sum_{r \in F} a_{r,j} = 1
    \right)
\end{equation}

\noindent\textbullet\ If node $j$ is a leaf, then it has no feature assigned to it.
\begin{equation}\label{eq:c11_leaves_no_feat}
    \bigwedge_{j \in N}
    \left(
        v_j \rightarrow \sum_{r \in F} a_{r,i} = 0
    \right)
\end{equation}

\noindent\textbullet\ For every positive (negative) example, if node $j$ is a leaf and it is assigned to class 0 (1), then at least one feature is discriminated with the value in the example at node $j$.
\begin{align}\label{eq:c12_13_row_discriminated}
    \bigwedge_{j \in N, e_q \in \varepsilon^{+}}
    \left(
        (v_j \land \neg c_j)
        \rightarrow
        \bigvee\limits_{r \in F} d_{r,j}^{\sigma(r,q)}
    \right) \\
    \bigwedge_{j \in N, e_q \in \varepsilon^{-}}
    \left(
        (v_j \land c_j)
        \rightarrow
        \bigvee\limits_{r \in F} d_{r,j}^{\sigma(r,q)}
    \right)\label{eq:c13}
\end{align}

\textbf{Additional Pruning hard clauses:}
Depending on which nodes are decided to be leaves or not, the possible list of children for a given node may change. These constraints enforce this simplification. We refer to the original description \cite{DBLP:conf/ijcai/NarodytskaIPM18}. We incorporate as in previous constraints the $\eta(i)$ variable to take into account in counters $\tau_{t,i}$ and $\lambda_{t,i}$ only the nodes that are used in the solution.

\begin{align}\label{eq:c_lambda_tau}
    \bigwedge_{i \in N} & \lambda_{0,i} \land \tau_{0,i} \\
    \bigwedge_{i \in N, t \in [1, \color{red}{\ceil{\frac{i}{2}}}\color{black}{}]} & 
    (
        \lambda_{t,i}
        \leftrightarrow
        (
            \lambda_{t,i-1} \lor \lambda_{t-1,i-1} \land v_i \color{red}{\land \eta(i)}\color{black}{}
        )
    ) \\
    \bigwedge_{i \in N, t \in [1, i]} & 
    (
        \tau_{t,i}
        \leftrightarrow
        (
            \tau_{t,i-1} \lor \tau_{t-1,i-1} \land \neg v_i \color{red}{\land \eta(i)}\color{black}{}
        )
    ) \\
    \bigwedge_{i \in N, t \in [1, \color{red}{\ceil{\frac{i}{2}}}\color{black}{}]} & \lambda_{t,i} \rightarrow \neg l_{i, 2(i-t+1)} \land \neg r_{i, 2(i-t+1) + 1} \\
    \bigwedge_{i \in N, t \in [\ceil{\frac{i}{2}}, i]} & \tau_{t,i} \rightarrow \neg l_{i, 2(t - 1)} \land \neg r_{i, 2t - 1}
\end{align}

\subsection{Multilabel encoding}
We extend \emph{MinDT} to support multilabels where $c_{k,j}$ variables now represent that the leaf $j$ belongs to class $k \in Cls$.
\begin{equation}
    \color{red}{}
    \bigwedge_{k \in Cls}
    \bigwedge_{j \in N, e_q \in \varepsilon^{k}}
    \left(
        (v_j \land c_{k,j})
        \rightarrow
        \bigvee\limits_{r \in K} d_{r,j}^{\sigma(r,q)}
    \right)
\end{equation}

\noindent\textbullet\ A decision node has no class assigned:
\begin{equation}
    \color{red}{}
    \bigwedge_{j \in N, k \in Cls} (\neg v_j \rightarrow \neg c_{k,j})
\end{equation}

\noindent\textbullet\ A leaf has only one label:
\begin{equation}
    \color{red}{}
    \bigwedge_{j \in N}\left( v_j \rightarrow \sum_{k \in Cls} = 1\right)
\end{equation}

\subsection{Partial MaxSAT encoding}

We assume a dataset $\varepsilon$, an upper ($ub$) and lower ($lb$) bounds on the size of the tree. Then:

\begin{proposition}
Let $PMSat_{PDT}^{\varepsilon,n=ub,lb}$ be $SoftN \wedge (eq.1) \wedge \ldots \wedge (eq.26)$.
The optimal cost of the Partial MaxSAT instance $PMSat_{PDT}^{\varepsilon,n=ub,lb}$ is
$\ceil{\frac{\lvert MPDT(\varepsilon)\rvert}{2}}$.
\end{proposition}

%%%%%%%%%%%%%%%%%%%%%%%%%%%%%%%%%%%%%%%%%%%%%%%%%%%%%%
%%%%%%%%%%%%%%%%%%%%%%%%%%%%%%%%%%%%%%%%%%%%%%%%%%%%%%

\section{The \emph{Linear} MaxSAT algorithm}\label{sec:linear-maxsat}
We present an slightly variation of the Linear SAT-based MaxSAT algorithm \cite{een_translating_2006,le_berre_sat4j_2010} that we will use to solve the Partial MaxSAT encoding presented in Section~\ref{sec:mindt_ms}.

Essentially, Algorithm \ref{alg:linear} solves a (Weighted) Partial MaxSAT optimization problem through a sequence of SAT queries. It receives as input parameters a (Weighted) Partial MaxSAT formula $\phi$ (a set of \emph{soft} $\phi_s$ and \emph{hard} $\phi_h$ clauses), an optional timeout $t$, and an optional upper bound. First, we create an incremental SAT solver $s$ (line~\ref{linear:create-sat}). Then, we add all the hard clauses $\phi_h$ and a copy of the soft clauses each extended with a new \emph{blocking} variable $b_i$ (line~\ref{linear:add-clauses}). 
Then, the upper bound $ub$ of $\phi$ is computed as the sum of the weights $w_i$ in $\phi_s$ plus one (line~\ref{linear:ub-init}). 
An \emph{incremental} PB encoder (an encoder which its $k$ can be updated) is initialized in line~\ref{linear:pb-init}. This PB encoder initially restricts the sum of the $b_i$ variables and their associated weights $w_i$ to be $\le ub$. In line \ref{linear:add-init-pb}, we use the PB encoder to translate into CNF the initial PB constraint. While this PB constraint in combination with the hard constraints is satisfiable, we extract the model, update the $ub$ to the cost of this model and refine the PB constraint (lines~\ref{linear:init-loop}~-~\ref{linear:end-loop}). Finally, we return the upper bound $ub$, the model $m$, and a reference to the incremental SAT solver $s$.

\input{algorithms/linear.tex}

%%%%%%%%%%%%%%%%%%%%%%%%%%%%%%%%%%%%%%%%%%%%%%%%%%%%%%
%%%%%%%%%%%%%%%%%%%%%%%%%%%%%%%%%%%%%%%%%%%%%%%%%%%%%%

\section{Extracting Multiple Diverse Solutions} \label{sec:mpdt-multiple}

We can obtain multiple solutions of a given optimal cost $c$ by generating the SAT instance representing the query of whether there is a solution of cost $c$. Then, we execute a SAT solver in incremental mode on the SAT instance, and whenever we get a solution, we add its negation to the SAT solver and ask the solver to solve the augmented instance.

In our approach, we take advantage of the fact that the SAT instance representing the optimal solutions is already created during the execution of the Linear MaxSAT algorithm which may be also augmented by useful learned clauses added in previous calls to the SAT solver. In particular, it is the last satisfiable SAT instance seen by Linear MaxSAT. 

To be able to access to the clauses of this last satisfiable instance we modify the Linear MaxSAT algorithms described in Algorithm~\ref{alg:linear} as follows. First, we move the statement of line~\ref{linear:pb-update} ($s.add\_clauses(pb.update(ub-1))$) inside the conditional block ($is\_sat = True$) of line~\ref{linear:if-sat} and add as an assumption to the solver in line~\ref{linear:solve} the auxiliary variables that reify the clauses in $pb.update(ub-1)$ \footnote{If $pb.update(ub-1)$ is a unit clause, we just add its literal as an assumption}. At this point, the formula containing all the possible optimal solutions corresponds to $s.clauses()$.

However, these solutions may be too \emph{similar}. In our case, we expect that diverse solutions (diverse MPDTs) may help to get a more robust approach in terms of accuracy. 

To enforce this diversity we will solve another MaxSAT problem. Algorithm MDSOL (Alg.~\ref{alg:multisol}) shows the pseudocode to extract multiple diverse solutions. As input we have the SAT formula that compiles the solutions, the number of solutions we want to extract, and the \emph{target} vars on which we will enforce our diversity criterion.

\input{algorithms/multisol}

The set of hard clauses of the MaxSAT formula we iteratively solve contains the clauses of the input SAT instance and the clauses discarding the solutions extracted so far (line~\ref{msol:literals}) restricted to the target variables. 

The soft clauses contain, as unit clauses of cost 1, the negation of the literals (restricted to the target variables) appearing in the solutions retrieved so far (line~\ref{msol:soft}). This way we prefer solutions where the polarity of the target variables appears less in the previous retrieved solutions. Notice that eventually, this approach can return all the optimal solutions.

%%%%%%%%%%%%%%%%%%%%%%%%%%%%%%%%%%%%%%%%%%%%%%%%%%%%%%
%%%%%%%%%%%%%%%%%%%%%%%%%%%%%%%%%%%%%%%%%%%%%%%%%%%%%%

\section{Computing Minimum Pure Decision Trees}

Finally, algorithm MPDT (Alg.\ref{alg:pdt-sampler}) describes how we compute MPDTs for a given dataset. We split randomly the input dataset into training and selection datasets, where $p$ is the percentage for examples to add to the training set (line~\ref{pdts:split}). 

In line~\ref{pdts:bounds} we perform a local search on the training dataset to find an upper bound on the tree size. In particular, we use ITI~\cite{DBLP:journals/ml/Utgoff89} (see Section~\ref{sec:experiments_times}) to find a pure decision tree (that is not guaranteed to be minimum).

Then, we create the MaxSAT encoding for computing MPDTs, described in Section~\ref{sec:mindt_ms} (line~\ref{pdts:encode}), on the training set $\varepsilon_{tr}$ and the lower and upper bounds.

In line~\ref{pdts:solve} we call the MaxSAT solver. We assume that as in the Linear MaxSAT pseudocode the MaxSAT solver returns a container for the SAT clauses compiling all the optimal solutions.

In line~\ref{pdts:multisol} we call MDSOL (Alg.~\ref{alg:multisol}) that extracts $k$ multiple \emph{diverse} MPDTs. We add as target vars the $a_{r,j}$ vars which represent that feature $r$ is assigned to node $j$. Preliminary experiments showed that using this variable is central and that adding other variables such as $u_{r,j}$ or $v_i$ does not lead to better results.

We consider also a variation, MPDT\_{lay}, inspired on \cite{DBLP:conf/sat/JanotaM20}. In short, we identify the $k'$ tree layouts to a certain depth that can lead to an MPDT. Then, we compute for each of these $k'$ layouts $k/k'$ solutions through MDSOL algorithm. 

From lines~\ref{pdts:decode}~to~\ref{pdts:update}, we add to $best_{dts}$ the MPDTs such that their accuracy, evaluated on the selection set $\varepsilon_{sel}$  is greater than or equal the best accuracy in $best_{dts}$ minus a deviation percentage $\delta$ (line~\ref{pdts:update}).
Finally, we return the set $best_{dts}$. In this paper, we just take randomly one of the MPDTs from $best_{dts}$ but other solutions combining more than one MPDT can be applied as other well-known approaches in machine learning (e.g. random forests).

\input{algorithms/mpdt-algorithm.tex}

%%%%%%%%%%%%%%%%%%%%%%%%%%%%%%%%%%%%%%%%%%%%%%%%%%%%%%
%%%%%%%%%%%%%%%%%%%%%%%%%%%%%%%%%%%%%%%%%%%%%%%%%%%%%%

\section{Preprocesing Datasets}

We use the \verb|KBinsDiscretizer| transformer from sklearn to convert the continuous variables into discrete values using 8 bins (intervals of values) with the uniform strategy (same amount of values for each range). Then, we convert all the non-binary features to binary using our custom binary encoding that applies a mapping from the original domain to a numeric domain. If the feature is ordinal, the order is kept. Then, those numeric values are represented in binary, and a new feature is created for each bit required to encode the entire domain for the original feature. Thus, for a feature with a domain $D$ we replace it with $log_2(\lvert D \rvert)$ new features.
As described in \cite{DBLP:conf/ijcai/NarodytskaIPM18}, the size of the encoding grows in $\mathcal{O}(kn^2+mnk)$, with $k$ being the number of features. Note that the binary encoding should have a natural advantage in the encoding size compared to the typical one-hot encoding (OHE) \footnote{The OHE encoding leads to worse results.} for non-binary datasets (see Table~\ref{tab:datasets}).

Finally, for each dataset benchmark, we split (following the stratification strategy) the dataset into a 64\% training set, 16\% selection set, and 20\% test set.

\section{Experimental Results}\label{sec:experiments_times}

As benchmarks we use datasets (binary and non-binary) from the CP4IM repository \cite{CP4IM}, the UCI repository \cite{Dua:2019}, and the PennML repository \cite{romano2021pmlb,Olson2017PMLB}. Table~\ref{tab:datasets} shows their statistics.
Our execution environment consists of a computer cluster with nodes equipped with two AMD 7403 processors (24 cores at 2.8 GHz) and 21 GB of RAM per core.

\input{tables/datasets}

\subsection{On MPDT runtime}
The first question we  address is whether the MPDT MaxSAT encoding described in Section~\ref{sec:mindt_ms} in combination with a MaxSAT solver allows us to compute faster MPDTs than the \emph{MinDT} approach described in \cite{DBLP:conf/ijcai/NarodytskaIPM18}.

For this, we adapted MinDT and the Linear algorithm (Algorithm~\ref{alg:linear}) used to solve the MaxSAT encoding to report the intermediate results found during the search process. Both of them are set to use the SAT solver CaDiCaL~\cite{Biere-SAT-Race-2019-solvers} as the SAT oracle. Then, we compare those approaches between them and against the best incomplete solvers on the unweighted track of the 2021 MaxSAT evaluation~\cite{bacchus2021maxsat}. In particular, we use the incomplete solvers \textit{SATLike-c}~\cite{lei2021satlike}, \textit{SATLike-ck}~\cite{lei2021satlike}, and \textit{TT-Open-WBO-Inc-21}~\cite{ttopenwbo21}.

We set a time limit of 1h and a memory limit of 20GB. We focused on large datasets (rows/features) and took 20 samples at each percentage $S_{pctgs}=\{40\%, 60\%, 80\%\}$, for a total of 580 instances for each size. This is a relevant experiment since if one wants to scale to large datasets as in \cite{Schidler2021SATbasedDT} using a kind of Large Neighborhood Search approach, it needs to efficiently solve samples of the original dataset with a SAT-based approach. Given these limits, we are interested in which approach is able to report the best upper bound (i.e. the smallest pure tree) on the given instances. For completeness, we also run the same experiment on the $80\%$ samples with a timeout of 10h.

The $ub$ for each instance is computed using the same approach as \cite{DBLP:conf/ijcai/NarodytskaIPM18}, i.e. the upper bound is the size of the computed $PDT(\varepsilon_{spl})$ using ITI algorithm \cite{DBLP:journals/ml/Utgoff89}. The $lb$ is trivially set to $3$.

Table~\ref{tab:incomplete} reports, for each sample size (column ``Sample size''), the number of instances (column ``\# Best UB'') where each approach (column ``Solver'') is able to report the best upper bound on the instance. Column ``Encoding'' indicates which encoding is used. 

\input{tables/incomplete}

As we can see, the \textit{satlike-ck} solver is the best solver, while \textit{MinDT} performs the worst in this experiment. This indicates that using the MaxSAT encoding allows for a faster reduction of the bounds in the MPDT problem.

Notice that MaxSAT algorithms share learnt clauses during the entire optimization process, do not require generating the SAT query at each step, and can skip suboptimal bounds. This is not the case for SAT incremental approaches as \textit{MinDT}, which can explain the differences that we observe in the results.

This experiment confirms it is useful to solve optimization problems as MaxSAT instances rather than with an agnostic SAT incremental approach provided memory requirements hold.

Notice anyway that the main goal of our paper is addressed in the next experiments, where most of the runtime is consumed on extracting multiple solutions rather than on computing the first MPDT.

\subsection{On MPDT accuracy}

The second question we want to address is how accurate can be the MPDT approach described in Section~\ref{sec:mpdt-multiple}. We compare with the DecisionTreeClassifier model from sklearn \cite{scikit-learndevelopersDecisionTreesScikit}. For these experiments, we set a time limit of 1 day and 100GB of memory.

We recall that datasets are split into 64\% training set, 16\% selection set, and 20\% test set. We try exactly 50 different random splits of the selection and test sets, select the best DT (from those returned by the particular approach) on the selection split and report the average accuracy on the test set.

For MPDT we evaluate \emph{mptd-1} that only generates one solution, and the variations that extract multiple solutions (with no diversity enforcement) (\emph{mpdt-nd}), multiple diverse solutions (\emph{mpdt}). Following \cite{DBLP:conf/sat/JanotaM20}, we also adapted our algorithm to consider the explicit \textit{layouts} for the decision trees, having \emph{mpdt-lay-nd} for the version that does not enforce diversity, and \emph{mpdt-lay} for the diverse solutions algorithm. This experiment considers the complete version of the MPDT algorithm.

For sklearn, we evaluate three variations on the training set with as many seeds as DTs returned by the MPDT version with more solutions. In pure sklearn (\emph{psk}) we run sklearn till it achieves 100\% accuracy.  In limited sklearn (\emph{lsk}) we restrict the DT size to the one reported by MPDT. Finally, in All sklearn (\emph{ask}) we explore each possible DT size from 3 to the maximum size reached by the \emph{psk} version. Notice that in \emph{ask} we evaluate many more DTs than for the previous approaches. Notice that \emph{lsk} and \emph{ask} may not compute Pure DTs in contrast to MPDT and \emph{psk}.

\newcommand{\h}[2]{\multicolumn{#1}{c}{#2}}

\begin{sidewaystable}[ht]
    \centering
    \small
\begin{tabular}{l|rrrrr|r|rrrrrrrrr}
\toprule
dataset & \h{2}{psk}    & \h{1}{lsk} & \multicolumn{2}{c|}{ask}   & \multicolumn{1}{c|}{opt} & \h{2}{mpdt-nd}  & \h{2}{mpdt-lay-nd} & \h{1}{mpdt-1}   & \h{2}{mpdt}     & \h{2}{mpdt-lay} \\
{}      &  t.acc & size &     t.acc  & t.acc & size &            &  \#sols & t.acc &   \#sols & t.acc & t.acc           & \#sols &  t.acc & \#sols &  t.acc   \\
\midrule
append  &  70.48 &   27 &      66.29 & 65.43 & 13   & 21         &   984   & 70.00 &     1167 & 70.00 & 70.00           &    592 &  \textbf{71.05} &    984 &  70.00  \\
audio   &  92.45 &   17 &      92.45 & 93.35 &  7   & 17         & 30000   & 93.95 &    30000 & 92.77 & \textbf{94.27}           &     48 &  93.50 &   4685 &  93.36  \\
back    &  69.67 &   33 &      74.28 & \textbf{74.44} &  3   & 25         &  5714   & 70.50 &      880 & 70.44 & 69.83           &      8 &  72.00 &     39 &  71.61  \\
corral  &    100 &   27 &      87.38 &   100 & 27   & \textbf{13}         &    20   & 99.25 &       20 & 99.25 &   \textbf{100}           &     20 &  99.25 &     20 &  99.25  \\
haber   &  80.26 &   73 &      77.33 & 79.51 & 25   & 47         &   624   & 79.00 &     1173 & 79.00 & \textbf{80.74}           &    216 &  79.00 &    624 &  79.00  \\
hayes-r &  79.33 &   41 &      77.00 & 77.08 & 31   & 35         &  1195   & \textbf{83.25} &     1722 & \textbf{83.25} & 78.08           &    349 &  82.92 &   1195 &  \textbf{83.25}  \\
hepa    &  60.84 &   27 &      60.84 & 64.83 & 17   & 23         &  1695   & \textbf{67.16} &       86 & 65.87 & 62.32           &      7 &  64.77 &     18 &  62.77  \\
h-votes &  93.31 &   41 &      94.00 & 93.98 & 15   & 33         & 11522   & \textbf{94.34} &    21007 & \textbf{94.34} & 94.25           &    370 &  93.56 &  11085 &  \textbf{94.34}  \\
iris    &  94.53 &   27 &      \textbf{94.53} & 91.46 & 19   & \textbf{25}         & 30000   & 94.53 &    30000 & 94.53 & 92.80           &    436 &  94.53 &  28280 &  94.53  \\
irish   &    100 &   15 &      97.94 & 99.28 & 13   &  \textbf{9}         &     6   &   \textbf{100} &        6 &   \textbf{100} &   \textbf{100}           &      6 &    \textbf{100} &      6 &    \textbf{100}  \\
lympho  &  \textbf{72.93} &   37 &      70.87 & 71.07 & 37   & 27         &  4214   & 71.73 &     5213 & 71.73 & 70.07           &    455 &  71.80 &   4214 &  71.73  \\
monks-1 &    \textbf{100} &   29 &        \textbf{100} &   \textbf{100} & 29   & 29         &  2368   &   \textbf{100} &     2496 &   \textbf{100} &   \textbf{100}           &    423 &    \textbf{100} &   2368 &    \textbf{100}  \\
monks-3 &  98.25 &   45 &      98.22 & \textbf{99.45} & 15   & 41         & 29952   & 99.11 &    30000 & 99.11 & 97.44           &    296 &  99.11 &   9050 &  99.11  \\
mush    & 100 &   23 &      99.97 & 99.97 & 23   & \textbf{15}         &   444   &   \textbf{100} &      477 &   \textbf{100} &   \textbf{100}           &    444 &    \textbf{100} &    444 &    \textbf{100}  \\
mux6    &  90.85 &   37 &      58.69 & 88.53 & 37   & \textbf{15}         &     2   &   \textbf{100} &        2 &   \textbf{100} &   \textbf{100}           &      2 &    \textbf{100} &      2 &    \textbf{100}  \\
new-thy &  92.60 &   33 &      \textbf{92.60} & 92.14 & 31   & \textbf{29}         & 30000   & 91.35 &    30000 & 91.35 & 89.58           &    423 &  91.35 &  18584 &  91.35  \\
spect   &  77.37 &   57 &      76.98 & 71.06 &  3   & 35         &   288   & \textbf{87.53} &      369 & \textbf{87.53} & \textbf{87.53}           &    288 &  \textbf{87.53} &    288 &  \textbf{87.53}  \\
wine    &  90.28 &   31 &      \textbf{95.33} & 92.78 & 27   & 23         &    32   & 92.44 &       32 & 92.44 & 92.67           &     32 &  92.44 &     32 &  92.44  \\
zoo     &  91.24 &   17 &      91.24 & 91.24 & 17   & 17         &  9364   & 91.24 &     9885 & 91.24 & 89.76           &    740 &  \textbf{94.76} &   9364 &  91.24  \\
\midrule
mean    &  87.07 &      &      84.15 & 86.61 &      &            &         & 88.70 &          & 88.57 & 87.86           &        &  \textbf{88.82} &        &  88.50  \\
median  &  91.24 &      &      91.24 & 91.46 &      &            &         & 92.44 &          & 92.44 & 92.67           &        &  \textbf{93.50} &        &  92.44  \\
\bottomrule
\end{tabular}
\caption{Test accuracy score. Size of the best solution is shown for non-optimal methods.}
    \label{tab:acc-binary}
\end{sidewaystable}

Table~\ref{tab:acc-binary} shows the average accuracy for each benchmark. For \emph{psk} and \emph{ask} we show the size of the DT (which may not be optimal). Additionally, for each dataset, we highlight in bold the approach with the best accuracy (ties are broken by DT size).

As we can see, extracting multiple diverse solutions (\emph{mpdt}) helps to improve the average accuracy of the MPDT approach. Moreover, with respect to sklearn, \emph{mpdt} obtains an average improvement on accuracy of 1.75\%, 4.7\% and 2.2\% with respect to \emph{psk}, \emph{lsk} and \emph{ask}. 

Therefore, on the same size \emph{mpdt} shows a much better accuracy-interpretability tradeoff than sklearn. Moreover, it is remarkable that \emph{mpdt} using much fewer DTs than \emph{ask}, and using Pure DTs, it is yet able to get a better average result. It seems that minimizing the size of PDTs helps to mitigate the overfitting phenomena. In this sense, it is interesting to explore the interactions with other properties such as the robustness of DTs \cite{moshkovitz2021connecting}.

Additionally, it is clear that \emph{mpdt} misses some good solutions, in particular, the first one used by \emph{mpdt-1}. Therefore, we also studied, whether any approach was able to \emph{see} a DT with better accuracy on the test set but was not chosen with the selection set. 
In particular, \emph{mpdt} obtains an average of 91.31\%, \emph{psk} 89.31\%, \emph{lsk} 86.65\% and \emph{ask} 90.51\%.

We also compare with \emph{DT(max)} \cite{DBLP:conf/ijcai/Hu0HH20} \footnote{DT(max) is comparable to DL8.5 \cite{DBLP:conf/aaai/AglinNS20} which improves \cite{verhaegheLearningOptimalDecision2019}.} which uses the incomplete MaxSAT solver Loandra~\cite{berg2020loandra}. Although Loandra is (according to the results of the MaxSAT evaluation \cite{bacchus2021maxsat}) superior to the Linear MaxSAT solver we use, our approach \emph{mpdt} still reports better results than \emph{DT(max)} using, as in the original paper, depth limits of 2, 3 and 4 (see Section~\ref{sec:related_work}). In particular, we find improvements of 7.86\%, 4.37\%, and 1.69\% respect to \emph{DT(max)} limited on depths 2, 3, and 4 respectively.

%%%%%%%%%%%%%%%%%%%%%%%%%%%%%%%%%%%%%%%%%%%%%%%%%%%%%%
%%%%%%%%%%%%%%%%%%%%%%%%%%%%%%%%%%%%%%%%%%%%%%%%%%%%%%

\section{Conclusions and Future Work}\label{sec:conclusions}

We have presented an efficient approach for computing MPDTs providing new insights on the accuracy-interpretability tradeoff. We have shown that by computing MPDTs for the training set, we can also get in practice accurate DTs on the test set compared to DT heuristic approaches in ML. This is a curious result since technically we overfit on the training set. As future work, we will explore further the relation of the size of the tree and its pureness on the training set in terms of prediction accuracy on the test set. We will also explore how to produce better multiple optimal solutions from our MaxSAT approach and combine them effectively to construct other predictors.

\FloatBarrier
\bibliography{main}
\bibliographystyle{unsrtnat}
\end{document}

%% file: algorithms/linear.tex
\begin{algorithm}[ht]
    \caption{Linear MaxSAT algorithm \label{alg:linear} (LMS)}
    
    \textbf{Input}: (Weighted) Partial MaxSAT formula $\phi \equiv \phi_s \cup \phi_h$ %, SAT solver $s$, Timeout $t$
    , Timeout $t$ (defaults to $\infty$)
    
    \textbf{Output}: Best upper bound $ub$, Best model $m$, SAT solver $s$
    
    \begin{algorithmic}[1]
    
    \State $s \leftarrow SAT\_Solver()$\label{linear:create-sat}
    
    \State $s.add\_clauses(\phi_h \cup \{c_i \lor b_i\ \lvert\ (c_i, w_i) \in \phi_s\})$ \label{linear:add-clauses}
    
    \State $ub \leftarrow \sum_{(c_i, w_i) \in \phi_s} w_i + 1$ \label{linear:ub-init}
    
    \State $pb \leftarrow \sum_{(c_i, w_i) \in \phi_s} w_i \cdot b_i \le ub$ \label{linear:pb-init}
    
    \State $s.add\_clauses(pb.to\_cnf())$ \label{linear:add-init-pb}
    
    \State $is\_sat, m \leftarrow True, \emptyset$
    
    \While{$is\_sat = True$ \textbf{and} $ub > 0$ \textbf{and} within timeout $t$} \label{linear:init-loop}
        \State $s.add\_clauses(pb.update(ub - 1))$ \label{linear:pb-update}
        
        \State $is\_sat \leftarrow s.solve()$ \label{linear:solve}
        
        \If{$is\_sat = True$}\label{linear:if-sat}
            \State $m \leftarrow s.model()$
        
            \State $ub \leftarrow cost(m)$ \label{linear:end-loop}
        \EndIf
    \EndWhile
    
    \State \textbf{return} $ub, m, s$
    \end{algorithmic}
\end{algorithm}

%% file: algorithms/multisol.tex
% Linear retorna a més el SAT solver
% Aquest sat solver té les hard.
% INPUT: SOL + SAT SOLVER
% bucle que va cridant linear, agafa el sat solver, fa backtrack, afegeix la soft, i torna a cridar a linear

%solve():
%  trees = EMPTY_SET
%  while len(trees) < k:
%    sol = solve(maxsat, formula)
%    trees += get_tree(sol)
%    new_soft, new_hard = get_clauses(sol)
%    update(formula, new_soft, new_hard)
%    maxsat.backtrack()
%  return trees
%
\begin{algorithm}[ht]
    \caption{Multiple Diverse Solutions (MDSOL)\label{alg:multisol}}
    \textbf{Input}: SAT instance $\varphi$, Num solutions $k$, target vars $vars$
    
    \textbf{Output}: Set of solutions $sols$
    
    \begin{algorithmic}[1]
    
    \State $\phi_h,\phi_s, sols \leftarrow \varphi, \emptyset, \emptyset$
    %\STATE $\phi_s \leftarrow \emptyset$
    % Empty formula as it is inside the SAT solver
    %\STATE $sols \leftarrow \emptyset$ 
    
    %\STATE $dts \leftarrow \emptyset$
    
    \While{$|sols| < k$}
    
        \State $ub,m,\_ \leftarrow LMS(\phi_h \cup \phi_s)$
        
        \If{$ub = \sum_{(c_i, w_i) \in \phi_s} w_i + 1$}
            \State \textbf{return} $sols$
        \EndIf
        \State $sols \leftarrow sols \cup \{m\}$
        \State $lits \leftarrow \{ l \lvert\ l \in m(vars)\}$ \label{msol:literals}
        \State $\phi_h \leftarrow \phi_h \cup \{(\bigvee_{l \in lits} \neg l,\infty)\}$
        
        \State $\phi_s \leftarrow \phi_s \cup \{ (\neg l, 1) \lvert\ l \in lits \}$\label{msol:soft}
        
        %\STATE $m' \leftarrow filter\_literals(m)$
        
        %\STATE $hard\_clauses \leftarrow \{\{-lit\ :\ lit \in m'\}\}$
        
        %\STATE $soft\_clauses \leftarrow \{(\{-lit\}, 1)\ :\ lit \in m'\}$
        
        %\STATE $s.extend\_clauses(hard\_clauses, soft\_clauses)$
        
        %\STATE $ub, m, s \leftarrow linear(\phi_e, s)$
        % FALTARIA EL BACKTRACK???
        
        %\STATE $dts \leftarrow dts \cup get\_tree(s)$
    
    \EndWhile
    
    \State \textbf{return} $sols$
    
    \end{algorithmic}
\end{algorithm}

%% file: algorithms/mpdt-algorithm.tex
\begin{algorithm}[ht]
    \caption{Minimum Pure Decision Tree (MPDT) \label{alg:pdt-sampler}}
    
    \textbf{Input}: Dataset $\varepsilon$, MaxSAT solver $msat$, Split pctg $p$, \\ Num solutions $k$, Accuracy deviation $\delta$
    
    \textbf{Output}: Best sample-based MPDTs $best_{dts}$
    
    \begin{algorithmic}[1]
    
    \State $\varepsilon_{tr}, \varepsilon_{sel} \leftarrow$ split $\varepsilon$ randomly according to $p$ \label{pdts:split}
    
    \State $best_{dts} \leftarrow \emptyset$ \Comment{Pairs of decision trees and accuracies}
    
    \State $ub, lb \leftarrow get\_bounds(\varepsilon_{tr})$ \label{pdts:bounds}
    
    \State $\phi \leftarrow encode(\varepsilon_{tr}, ub, lb)$ \label{pdts:encode}
    
    \State $\_,\_,s \leftarrow msat(\phi)$\label{pdts:solve}
    
    \State $sols \leftarrow MDSOL(s.clauses,k,\{a_{r,j} \lvert a_{r,j} \in vars(\phi)\})$\label{pdts:multisol}

    \For{$sol \in sols$}
        
        \State $dt \leftarrow decode(sol)$\label{pdts:decode}
        
        \State $acc_{dt} \leftarrow evaluate(dt, \varepsilon_{sel})$ \label{pdts:acc-selection-start}
        
        \State $best_{dts} \leftarrow update(best_{dts}, dt, acc_{dt}, \delta)$\label{pdts:update}
    \EndFor
    
    \State \textbf{return} $best_{dts}$
    
    \end{algorithmic}
\end{algorithm}

%% file: tables/datasets.tex
\begin{table}[ht]
\centering
\begin{tabular}{lrrrrrrr}
\toprule
                 & \multicolumn{2}{c}{\#Columns} & \#Labels & \multicolumn{4}{c}{\#Rows} \\
                 &  binary &  ohe &   &   total &  train &  sel &  test \\
\midrule
% anneal-un        &           89 &        89 &         2 &    778 &         497 &       125 &        156 \\
append     &           22 &        56 &         2 &    105 &          67 &        17 &         21 \\
audio     &          146 &       146 &         2 &    216 &         137 &        35 &         44 \\
% australian       &           34 &        72 &         2 &    684 &         437 &       110 &        137 \\
back         &           50 &        89 &         2 &    180 &         115 &        29 &         36 \\
% balance-scale    &           13 &        21 &         3 &    625 &         400 &       100 &        125 \\
% balloons1        &            5 &         5 &         2 &     20 &          12 &         4 &          4 \\
% balloons2        &            5 &         5 &         2 &     20 &          12 &         4 &          4 \\
% balloons3        &            5 &         5 &         2 &     16 &           9 &         3 &          4 \\
% balloons4        &            5 &         5 &         2 &     20 &          12 &         4 &          4 \\
% bupa             &           16 &        39 &         2 &    283 &         180 &        46 &         57 \\
% cancer           &           37 &        90 &         2 &    683 &         436 &       110 &        137 \\ % TODO cita
% car              &           13 &        22 &         2 &   1728 &        1105 &       277 &        346 \\
% colic            &           80 &       416 &         2 &    368 &         235 &        59 &         74 \\
% connect-4        &           85 &       127 &         3 &  67557 &       43236 &     10809 &      13512 \\
% contraceptive    &           22 &        69 &         3 &   1406 &         899 &       225 &        282 \\
corral           &            7 &         7 &         2 &    160 &         102 &        26 &         32 \\
% ecoli            &           17 &        42 &         5 &    318 &         203 &        51 &         64 \\
% glass            &           28 &        67 &         5 &    193 &         123 &        31 &         39 \\
haber         &           11 &        28 &         2 &    266 &         169 &        43 &         54 \\
hayes-r       &            9 &        16 &         3 &    120 &          76 &        20 &         24 \\
% heart-c          &           30 &        61 &         2 &    303 &         193 &        49 &         61 \\
% heart-statlog    &           29 &        59 &         2 &    270 &         172 &        44 &         54 \\
hepa        &           58 &       321 &         2 &    155 &          99 &        25 &         31 \\
h-votes   &           17 &        17 &         2 &    435 &         278 &        70 &         87 \\
% hungarian        &           41 &       298 &         2 &    294 &         188 &        47 &         59 \\
% hypothyroid-un   &           84 &        84 &         2 &   3230 &        2067 &       517 &        646 \\
iris             &           13 &        32 &         3 &    150 &          96 &        24 &         30 \\
irish            &           16 &        53 &         2 &    500 &         320 &        80 &        100 \\
% kr-vs-kp         &           38 &        39 &         2 &   3196 &        2044 &       512 &        640 \\
% lenses           &            6 &         7 &         3 &     24 &          15 &         4 &          5 \\
% letter           &           49 &       129 &        26 &  19931 &       12755 &      3189 &       3987 \\
% lung\_cancer     &           19 &        73 &         2 &     59 &          37 &        10 &         12 \\ % TODO cita
lympho     &           30 &        51 &         4 &    148 &          94 &        24 &         30 \\
% meteo            &            7 &         9 &         2 &     14 &           8 &         3 &          3 \\ % TODO cita
monks-1          &           11 &        16 &         2 &    556 &         355 &        89 &        112 \\
% monks-2          &           11 &        16 &         2 &    601 &         384 &        96 &        121 \\
monks-3          &           11 &        16 &         2 &    548 &         350 &        88 &        110 \\
% mouse            &           21 &        46 &         2 &     70 &          44 &        12 &         14 \\ % TODO cita
mushr      &          112 &       112 &         2 &   8124 &        5199 &      1300 &       1625 \\
mux6             &            7 &         7 &         2 &    128 &          81 &        21 &         26 \\
new-thy      &           16 &        39 &         3 &    212 &         135 &        34 &         43 \\ % TODO cita
% person           &           15 &       172 &         6 &    500 &         320 &        80 &        100 \\
%primary-tumor-un &           32 &        32 &         2 &    321 &         204 &        52 &         65 \\
%promo        &          115 &       229 &         2 &    106 &          67 &        17 &         22 \\
% schizo           &           95 &      2155 &         2 &    340 &         217 &        55 &         68 \\
% shuttleM         &           61 &       692 &         2 &  14500 &        9280 &      2320 &       2900 \\
% soybean-un       &           51 &        51 &         2 &    628 &         401 &       101 &        126 \\
spect            &           23 &        23 &         2 &    251 &         160 &        40 &         51 \\
% splice-1-un      &          288 &       288 &         2 &   3189 &        2040 &       511 &        638 \\
% tae              &           12 &        27 &         3 &    134 &          85 &        22 &         27 \\
% tic-tac-toe-un   &           28 &        28 &         2 &    958 &         612 &       154 &        192 \\
wine             &           47 &       263 &         3 &    178 &         113 &        29 &         36 \\
zoo              &           26 &       122 &         7 &    101 &          64 &        16 &         21 \\
\bottomrule
\end{tabular}
\caption{Statistics from the datasets used in the experimental results. In order: the number of features for both the binary encoding and the OHE, the number of labels (2 for binary classification instances, more for multilabel), the total number of rows in the dataset, and the number of rows used as train, selection, and test.}
\label{tab:datasets}
\end{table}

%% file: tables/incomplete.tex
\begin{table*}
    \centering
\begin{tabular}{l|l|l|l|c}
\toprule

\makecell[cl]{Sample\\size} & Timelimit & Solver & Encoding & \# Best UB \\
\midrule
40\% & 1h & satlike-ck & MaxSAT & 528 \\
40\% & 1h & MPDT & MaxSAT & 505 \\
40\% & 1h & satlike-c & MaxSAT & 504 \\
40\% & 1h & TT-Open-WBO-Inc-21 & MaxSAT & 489 \\
40\% & 1h & MinDT & SAT & 469 \\
\midrule
60\% & 1h & satlike-ck & MaxSAT & 491 \\
60\% & 1h & satlike-c & MaxSAT & 426 \\
60\% & 1h & MPDT & MaxSAT & 421 \\
60\% & 1h & TT-Open-WBO-Inc-21 & MaxSAT & 417 \\
60\% & 1h & MinDT & SAT & 404 \\
\midrule
80\% & 1h & satlike-ck & MaxSAT & 422 \\
80\% & 1h & satlike-c & MaxSAT & 383 \\
80\% & 1h & MPDT & MaxSAT & 404 \\
80\% & 1h & TT-Open-WBO-Inc-21 & MaxSAT & 382 \\
80\% & 1h & MinDT & SAT & 374 \\
\midrule
80\% & 10h & satlike-ck & MaxSAT & 510 \\
80\% & 10h & satlike-c & MaxSAT & 405 \\
80\% & 10h & MPDT & MaxSAT & 425 \\
80\% & 10h & TT-Open-WBO-Inc-21 & MaxSAT & 406 \\
80\% & 10h & MinDT & SAT & 400 \\
\bottomrule
\end{tabular}
    \caption{Number of instances where the best upper bound was reported by each incomplete solver.}
    \label{tab:incomplete}
\end{table*}